\documentclass{article}
\usepackage{spconf,amsmath,graphicx}
\usepackage{caption}
\usepackage{subcaption}
\usepackage{wrapfig}
\usepackage{hyperref}
\usepackage{enumitem}

\title{Bounding Box Disparity: 3D Metrics for Object Detection\\ With Full Degree of Freedom}

\name{Michael G. Adam, Martin Piccolrovazzi, Sebastian Eger, Eckehard Steinbach\thanks{This work is funded by Germany's Federal Ministry of Education and Research within the project KIMaps (grant ID \#01IS20031C).}}
\address{Chair of Media Technology and Munich Institute of Robotics and Machine Intelligence,\\ Technical University of Munich, Munich, Germany}

\usepackage[absolute,overlay]{textpos}
\begin{document}
\begin{textblock*}{175.5truemm}(1.92cm,1.3cm)
\noindent\fbox{%
    \parbox{\textwidth}{%
\footnotesize \copyright \, 2022 IEEE. Personal use of this material is permitted. Permission from IEEE must be obtained for all other uses, in any current or future media, including reprinting/republishing this material for advertising or promotional purposes, creating new collective works, for resale or redistribution to servers or lists, or reuse of any copyrighted component of this work in other works.
    }%
}%
\end{textblock*}%
\maketitle

\begin{abstract}
  The most popular evaluation metric for object detection in 2D images is Intersection over Union (IoU). 
  Existing implementations of the IoU metric for 3D object detection usually neglect one or more degrees of freedom. In this paper, we first derive the analytic solution for three dimensional bounding boxes. 
  As a second contribution, a closed-form solution of the volume-to-volume distance is derived. 
  Finally, the Bounding Box Disparity is proposed as a combined positive continuous metric. We provide open source implementations of the three metrics as standalone python functions, as well as extensions to the Open3D library and as ROS nodes.
\end{abstract}
\vspace{-0.5em}
\section{Introduction}
\begin{figure}%
    \centering
    \begin{subfigure}{.5\linewidth}
        \centering
        \includegraphics[trim=375 200 270 250, clip, width=0.96\linewidth]{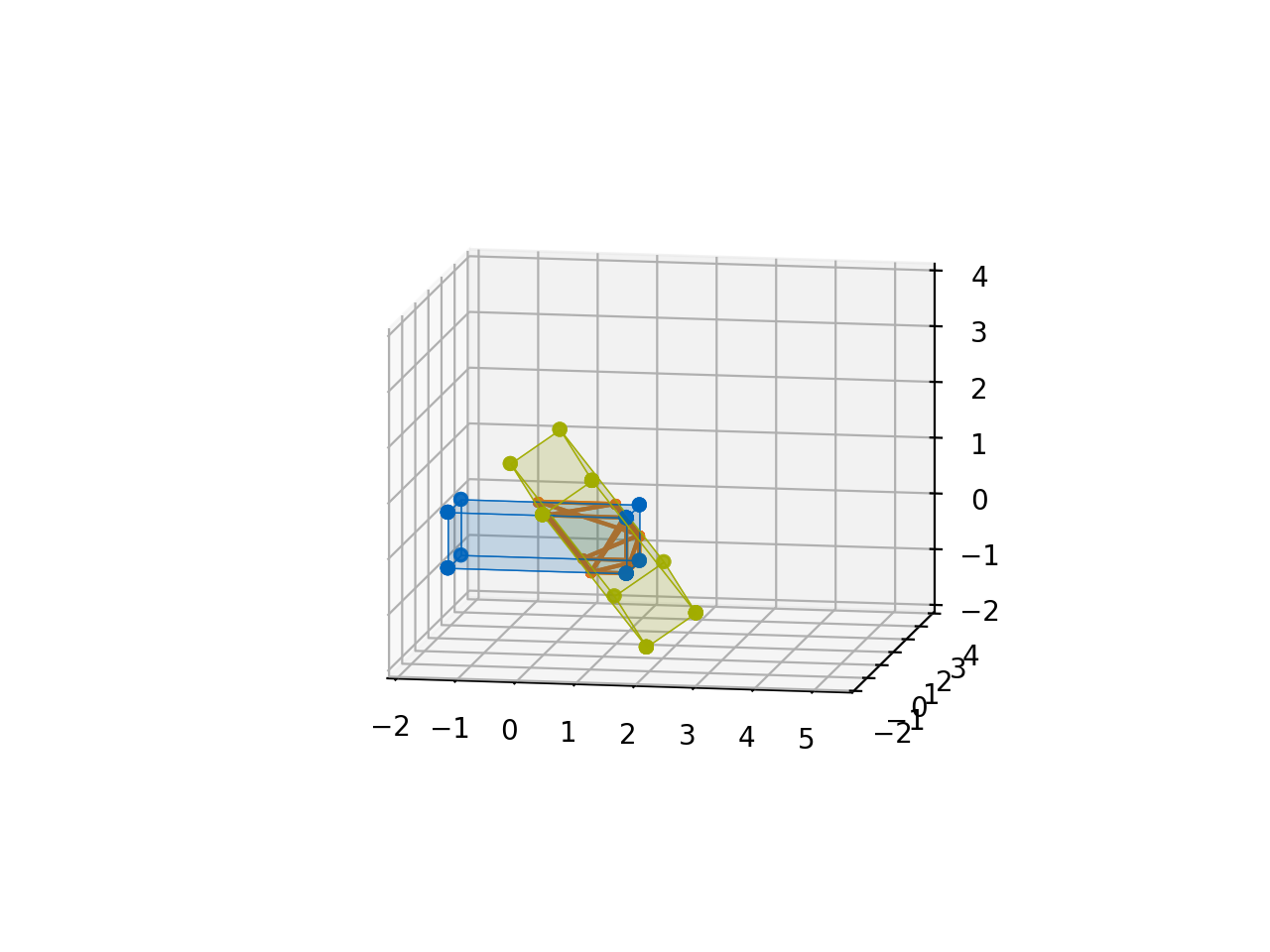}
        \caption{}
        \label{fig:sub1}
    \end{subfigure}%
    \begin{subfigure}{.5\linewidth}
        \centering
        \includegraphics[trim=375 200 270 250, clip, width=0.96\linewidth]{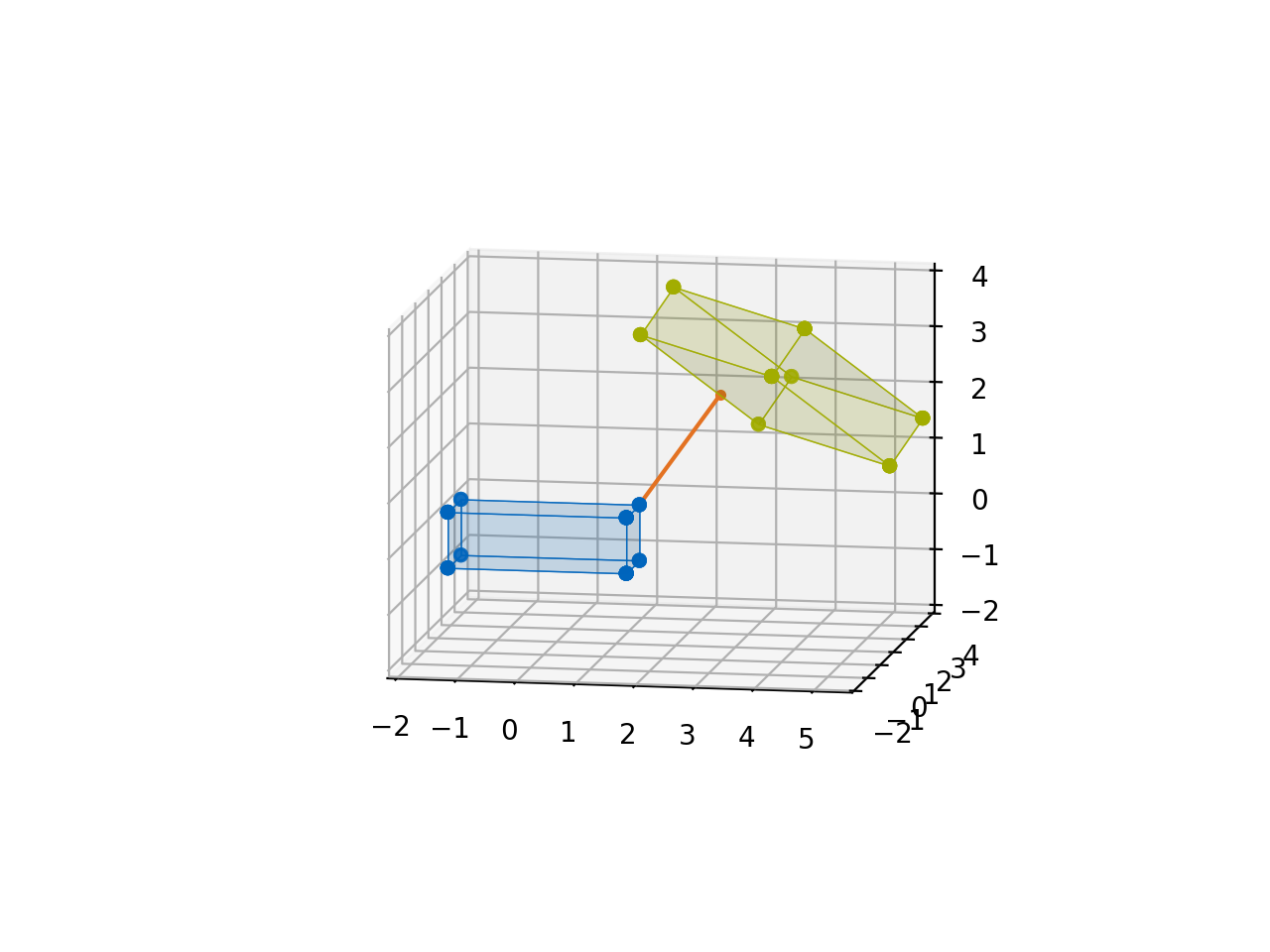}
        \caption{}
        \label{fig:sub2}
    \end{subfigure}
    \vspace{-0.8em}
    \caption{Similarity metrics for two 3D bounding boxes. \\$\; $(a) Intersection over Union (IoU) \\(b) volume-to-volume distance (v2v)}
    \label{fig:metrics}
    \vspace{-1.em}
\end{figure}

As 3D object detection gets more popular and new datasets are published \cite{data:matter,data:apple,data:google},
evaluation metrics gain in importance. The most common one is Intersection over Union (IoU). It is well known from object detection on two-dimensional data such as images. 
Existing implementations of its three-dimensional counterpart usually neglect one or more degrees of freedom. Examples for this are implementations that work with axis-aligned bounding boxes or only consider a rotation around the z-axis \cite{paper:3.5DIoU}. This oversimplifies real world problems, as usually rotation of objects is possible in any given direction.\\
To the best of our knowledge, although the strategy of computing IoU has already been mathematically generalized \cite{paper:generalization}, we provide and derive the first closed-form analytic solution for the case of 3D bounding boxes with full degree of freedom.\\
We further derive an analytic solution for the volume-to-volume distance (v2v) of two 3D bounding boxes. The metric v2v is defined as the shortest distance between the hull of one volume to the hull of another volume. Both metrics are visualized in Fig. \ref{fig:metrics}.\\
For both metrics we provide the first open source implementation as a standalone python function, as well as an extension to the Open3D library \cite{famew:o3d} and a ROS-node \cite{famew:ros}.\\
This paper is structured as follows. First, we discuss related work. In Section 3, we mathematically define bounding boxes and review the existing metrics. Afterwards the solution for volumetric IoU is stated and shortly compared to its point based method. Consecutively v2v is presented and a combined positive continuous metric called Bounding Box Disparity (BBD) is proposed.

\section{Related work}
Calculating the intersection of two three-dimensional volumes is a common task in rendering, CAD pipelines and deep learning \cite{paper:hands}. In open source frameworks such as blender~\cite{famew:blender} it is possible to use an accurate solver. In practice, fast numerical solutions are used, in order to speed up the computation with the disadvantage of loosing accuracy.  Instead of using cuboids directly, the frameworks usually have in common that only triangular-mesh representations are allowed for boolean operations such as intersection.\\
Hence, with an additional step of meshifying bounding boxes one can define a volumetric intersection. This adds unnecessary computation of multiple equations and conversions. For reference purposes, we also provide a blender-based implementation of this method.
When used for custom code this adds a big dependency, since blender and its API is rather developed for graphical usage with complex scenes, but not for calculating basic linear equations.\\
The Open3D framework, which is often used in the context of machine learning does not have an implementation of an exact intersection for meshes, as well as no solution for oriented bounding boxes.\\
Benchmarks for 3D object detection such as \cite{data:standford} often rely on evaluation metrics that use axis aligned bounding boxes or bounding boxes with only a rotation around the z-axis \cite{data:sunrgbd, data:scannet}. Other benchmarks like \cite{data:kitti} use a combination of 2D IoU and an additional orientation value.
To the best of our knowledge there are no full-degree of freedom bounding box benchmarks published for the 3D case yet.
\begin{figure}
    \centering
    \includegraphics[trim=0cm 12.06cm 26.4cm 0.1cm, clip, width=0.55\linewidth]{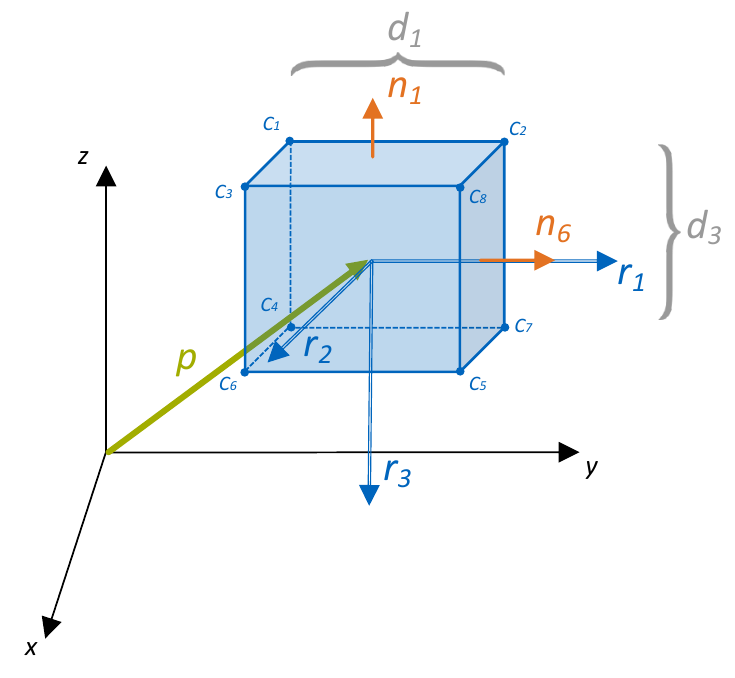}
    \caption{Bounding box definitions.}
    \vspace{-.6em}
    \label{fig:bb_def}
    \vspace{-.6em}
\end{figure}

\section{Methods}
\subsection{Definition of a Bounding Box}
First, we have to define a bounding box $i$, shown in Fig.~\ref{fig:bb_def}. It can be represented by a transformation matrix $T^i$, consisting of its rotation $R^i$, position $p^i$ and dimension $d^i$
\begin{equation}
    T^i = \left[ \begin{array}{c|c}
        \scalebox{1.2}{$d^i R^i$} & \scalebox{1.2}{$p^i$} \\
        \hline
        0\;0\;0 & 1
    \end{array} \right],
\end{equation}
whereas the position corresponds to the cuboid center. Its corners $C_{1-8}^i$ can then be defined as follows:
\begin{equation}
    C_{1-8}^i = T^i u_{1-8}
\end{equation}
with $u_{1-8}$ being defined as the corners of the unit cube centered around the origin of the coordinate system.
\begin{equation}
\scalebox{0.7}{%
    $u_{1-8} = \begin{bmatrix} 
        -0.5 & 0.5 & -0.5 & -0.5 & 0.5 & -0.5 & 0.5 & 0.5 \\ 
        -0.5 & -0.5 & 0.5 & -0.5 & 0.5 & 0.5 & -0.5 & 0.5 \\ 
        -0.5 & -0.5 & -0.5 & 0.5 & 0.5 & 0.5 & 0.5 & -0.5 
    \end{bmatrix}$
    }
\end{equation}
One can further define the edges $e_{1-12}^i$ and faces $f_{1-6}^i$ of a cube as a list of connected corners. Its elements correspond to the column-index in the matrix of corners.
\begin{align}
    e_{1-12} = &[[1, 2], [2, 8], [3, 8], [1, 3], [4, 7], [7, 5], \nonumber \\
                & [6, 5], [4, 6], [1, 4], [2, 7], [8, 5], [3, 6]] \\
    f_{1-12} = &[[1, 2, 3], [1, 2, 4], [1, 3, 4], \nonumber\\
                & [5, 6, 7], [5, 6, 8], [5, 7, 8]]
\end{align}
Lastly, the normal vectors $n_{1-6}^i$ of the faces are defined by the rotation $R^i$. Each column $r_{j}^i$ corresponds to two face normals~-~one time in positive and one time in negative direction.
\begin{align}
    R^i &= [r_1^i,r_2^i,r_3^i] \nonumber \\
    n_1^i &= - r_3^i \qquad n_2^i = - r_2^i \qquad n_3^i = - r_1^i \\
    n_4^i &= + r_3^i \qquad n_5^i = + r_2^i \qquad n_6^i = + r_1^i \nonumber
\end{align}

\subsection{Metrics}
There a several metrics to consider when comparing two bounding boxes (BB) \cite{paper:metrics,paper:metrics2}:
\begin{itemize}[leftmargin=1em]
	\itemsep-0.5em 
    \item absolute/quadratic difference in position
    \item absolute/quadratic difference in size
    \item rotation (can affect difference in size \cite{paper:orient}), i.e.:
    		\vspace{-0.5em}
        \subitem - angular-difference of euler angles
		\vspace{-0.5em}        
        \subitem - quaternion-distance
        \vspace{-0.5em}
        \subitem - distance of rotation matrices
    \item IoU of point-cloud points within the BBs \cite{paper:pointsIoU}
    \item volumetric IoU
    \item volume-to-volume distance
\end{itemize}
All of the above are part of our open source implementation. Only the last two are discussed in the following.

\subsection{Volumetric IoU}
\subsubsection{Points of interest}
We now define possible corner points $POI$ of the intersection. There are four candidates: 
\begin{itemize}[leftmargin=1em]
	\itemsep-0.5em 
    \item Corners of the first cuboid laying inside the second cuboid
    \item Corners of the second cuboid laying inside the first cuboid
    \item Intersections of an edge of the first cuboid with a plane of the second cuboid
    \item Intersections of an edge of the second cuboid with a plane of the first cuboid
\end{itemize}
In order to compute them, we formulate equations for each line and plane corresponding to the edges and faces of both cubes.
The lines can simply be defined as
\begin{equation}
    L_{k}^i \left(t_l\right) = e_{k,1}^i + \underbrace{\left( e_{k,2}^i - e_{k,1}^i\right)}_{\substack{m_k^i}} t_l
\end{equation}
with $m_k^i$ describing the line's slope.
Whereas the planes can be defined by three corner points.
\begin{align}
    P_{k}^i \left(t_p\right) = f_{k,1}^i + \underbrace{\begin{bmatrix}
        f_{k,2}^i - f_{k,1}^i & f_{k,3}^i - f_{k,1}^i
    \end{bmatrix}}_{\substack{N_k^i}} \begin{bmatrix}
        t_{p,1} \\
        t_{p,2}
    \end{bmatrix}
    \label{eq:plane}
\end{align}
The $3\times2$-matrix $N_k^i$ can also be expressed by the normals perpendicular to the face normal $n_k^i$ scaled according to $d^i$.\\
For all possible line-plane combinations of the two cuboids, we get $144$ equations. Each resulting in possible points of interest. The number depends on how many lines are parallel with the planes or lay in the planes. This again depends on how the cubes are rotated to each other.\\
Next, we calculate the corners $C^1,C^2$ of both cuboids and add them to the list of points. 

\subsubsection{Checking Validity of the Points}
Now every point in the solution list has to be checked, if its a valid corner point of the intersection. For this the points have to be part of both cubes.\\
This can be done by transforming the points into the coordinate system of $T^1$.
\begin{equation}
    POI = {T^1}^{-1} POI 
\end{equation}
The corner points of the first cube $C_1$ now correspond to the ones of a unit cube $u_{1-8}$. Thus, every point has to be checked if its coordinates are smaller than $0.5$ and greater than $-0.5$.
\begin{equation}
    valid = (POI < 0.5)\, \& \,(POI > -0.5)
    \label{eq:validity}
\end{equation}
After pruning the non valid points from the list, we transform everything into the coordinate system of the second cube $T^2$.
\begin{equation}
    POI = {T^2}^{-1} T^1 POI 
\end{equation}
Repeating the validity check from equation \ref{eq:validity} leaves us all relevant points.

\subsubsection{Calculating the Volume}
We now pick every valid point and construct the convex hull of this set of points, which corresponds to the intersection of the cuboids. If the hull cannot be constructed, because no point is valid or the points all lay in a plane, then there is no intersection, thus $IoU=0$ holds. \\
In all other cases, the volume of the hull $V_I$ can be computed by summation of the signed volumes of tetrahedrons \cite{math:signedV}. \\
As a last step we calculate the volume of the union $V_U$
\begin{equation}
    V_U = d_{1}^1 d_{2}^1 d_{3}^1 + d_{1}^2 d_{2}^2 d_{3}^2 - V_I
\end{equation}
such that $IoU$ is defined as
\begin{equation}
    IoU = \frac{V_I}{V_U}.
\end{equation}

\begin{figure}
    \centering
    \includegraphics[trim=0.55cm 7.22cm 12.68cm 0.6cm, clip, width=1.\linewidth]{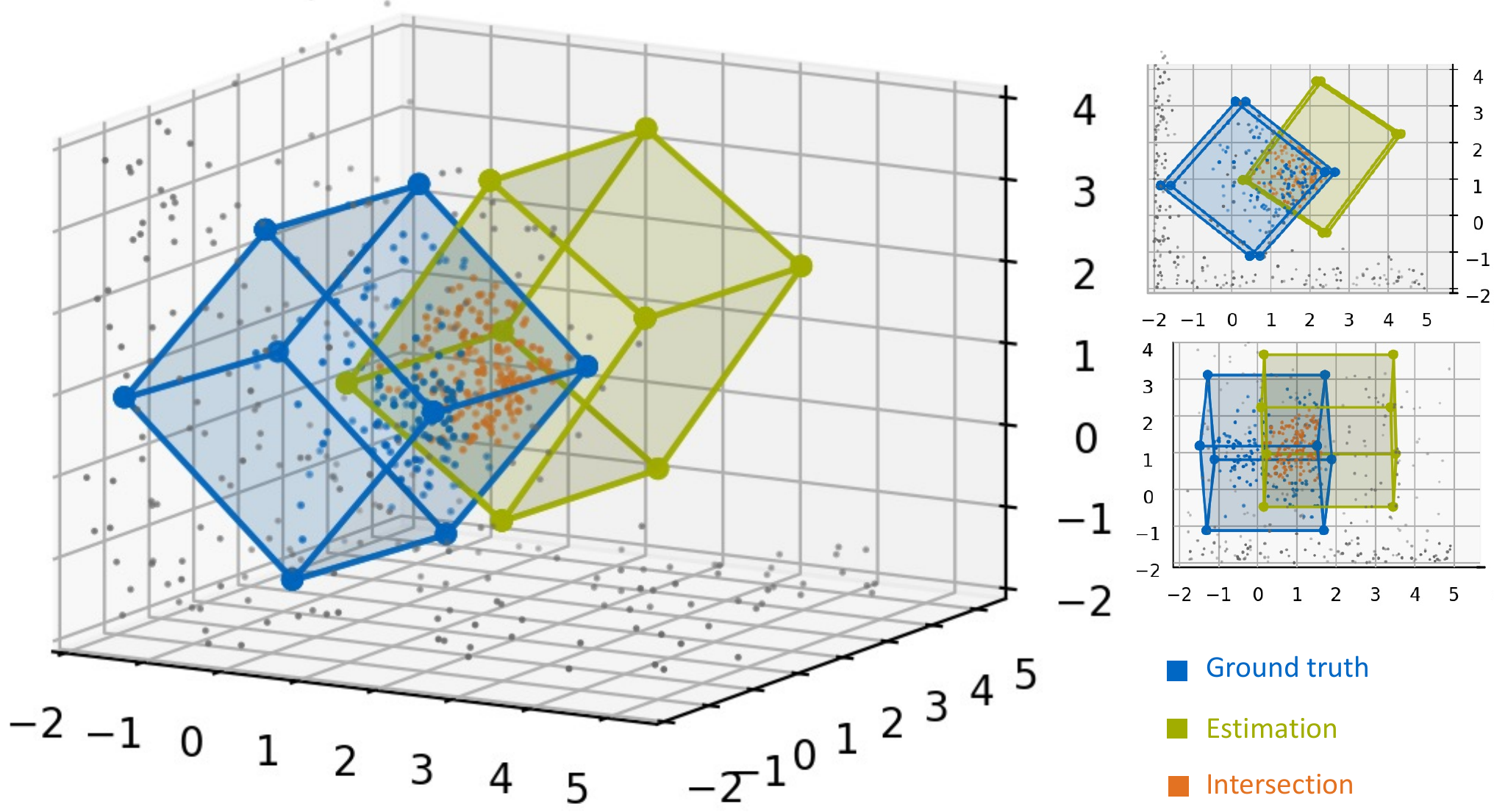}
    \caption{Object detection example with underlying point cloud.}
    \label{fig:example}
    \vspace{-0.8em}
\end{figure} 
\subsection{Comparison of point and volume based IoU}
In Fig~\ref{fig:example} a typical example of an object leaning against a wall is shown. The underlying point cloud is constructed 
by a lidar scan. Hence, most of the points of the object lay in the front. With the shown estimation of an object detector, this results in a relatively high intersection over union of $0.47$, when computed based on the points in the cloud, in comparison to $0.07$ of the semantically more meaningful volumetric counterpart.

\vspace{-0.5em}
\subsection{Volume-to-Volume Distance}
\subsubsection{Point-Pairs of Interest}
\begin{figure}[b]%
    \centering
    \begin{subfigure}{.25\linewidth}
        \centering
        \includegraphics[trim=375 200 270 250, clip, width=.9\linewidth]{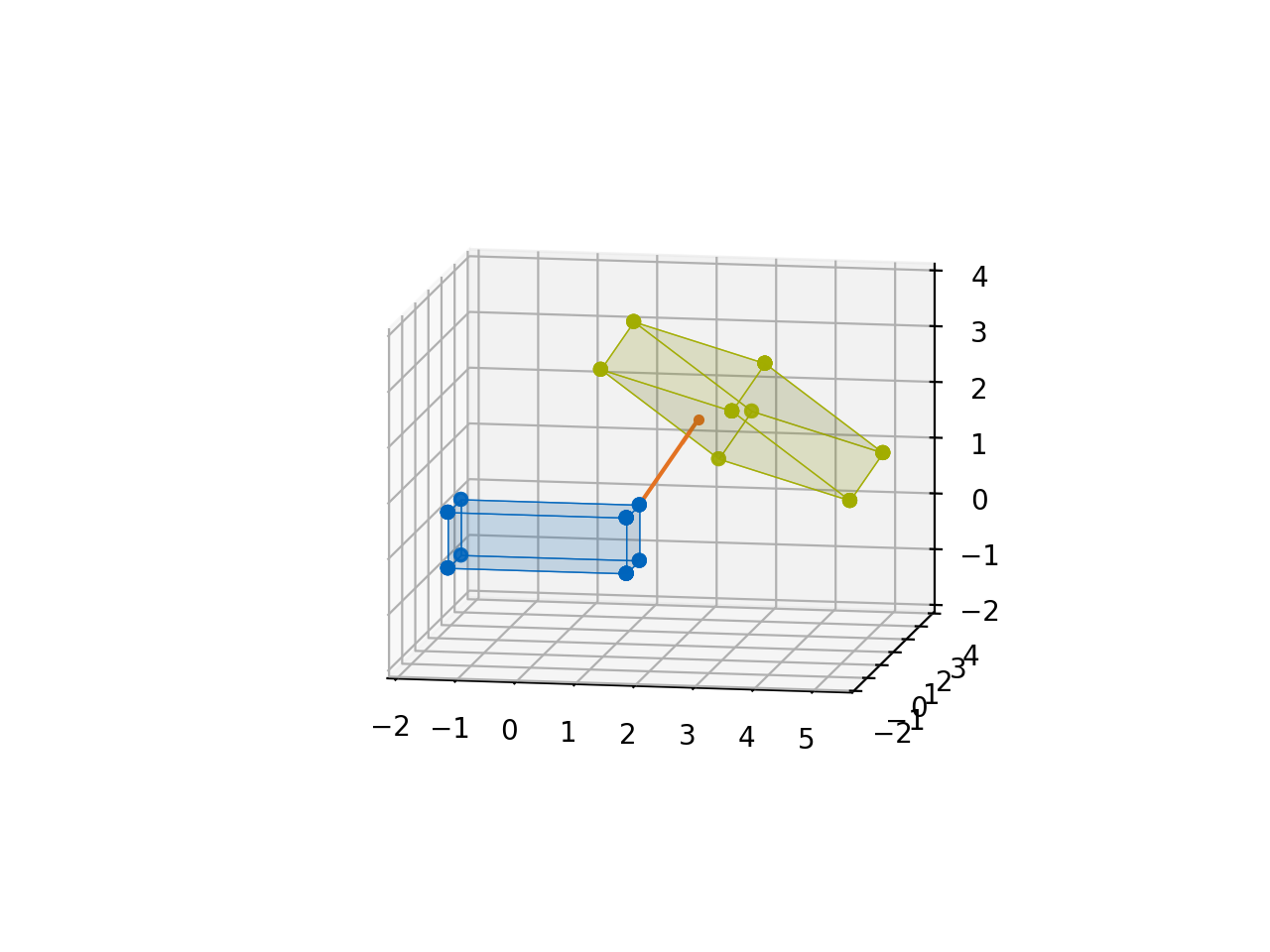}
        \caption{}
        \label{fig:c-f}
    \end{subfigure}%
    \begin{subfigure}{.25\linewidth}
        \centering
        \includegraphics[trim=375 200 270 250, clip, width=.9\linewidth]{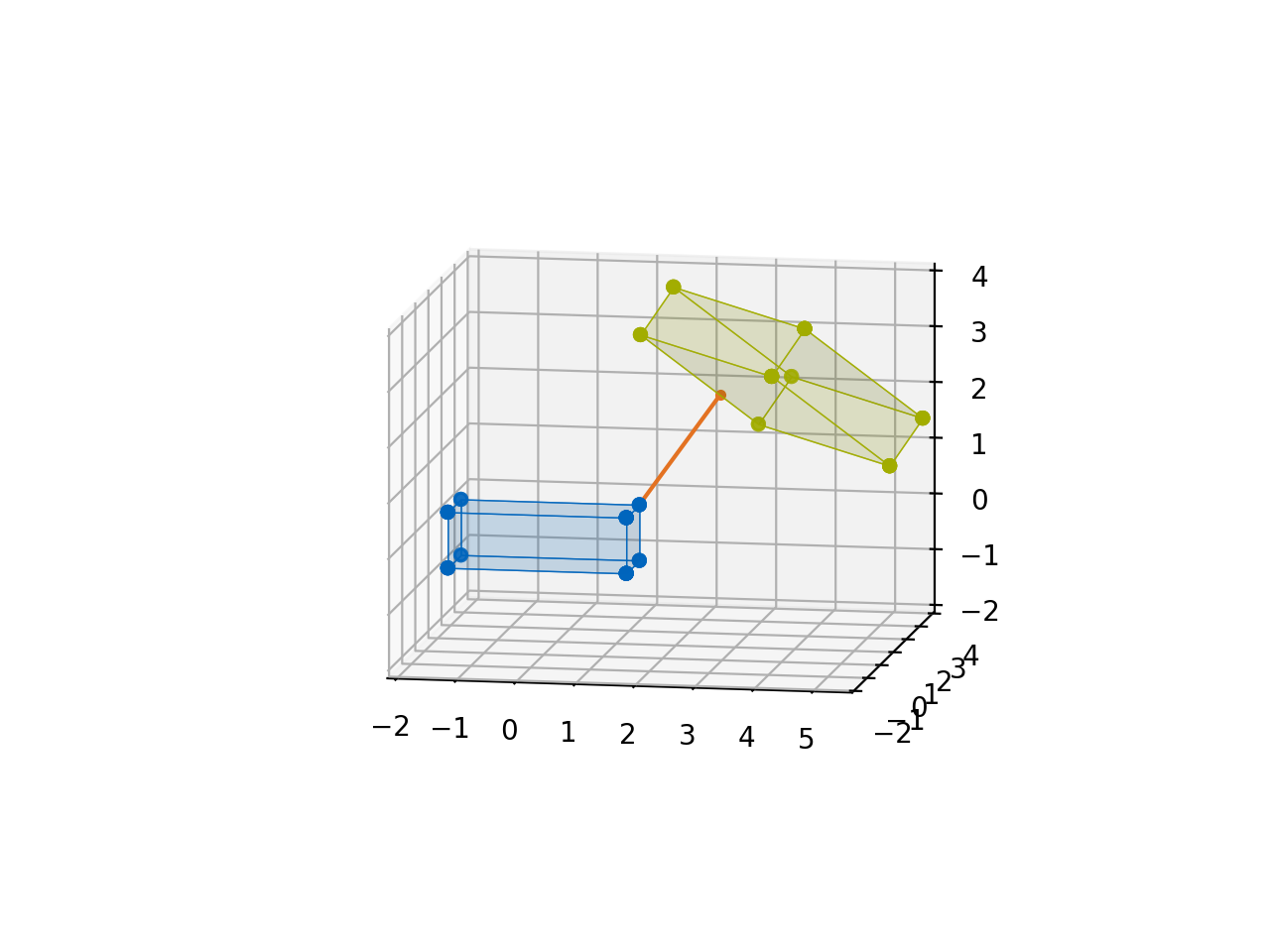}
        \caption{}
        \label{fig:c-e}
    \end{subfigure}%
     \begin{subfigure}{.25\linewidth}
        \centering
        \includegraphics[trim=375 200 270 250, clip, width=.9\linewidth]{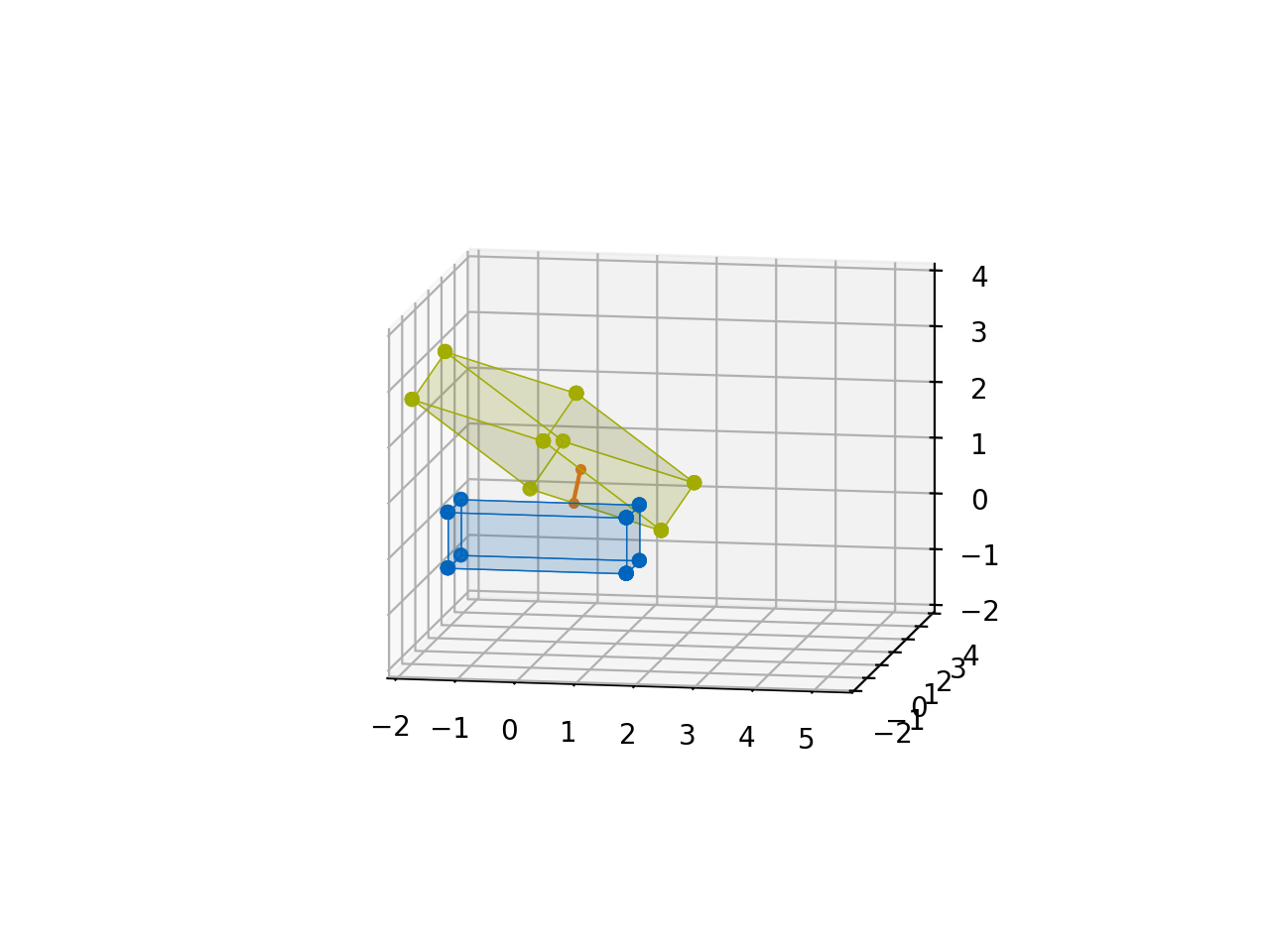}
        \caption{}
        \label{fig:e-e}
    \end{subfigure}%
    \begin{subfigure}{.25\linewidth}
        \centering
        \includegraphics[trim=375 200 270 250, clip, width=.9\linewidth]{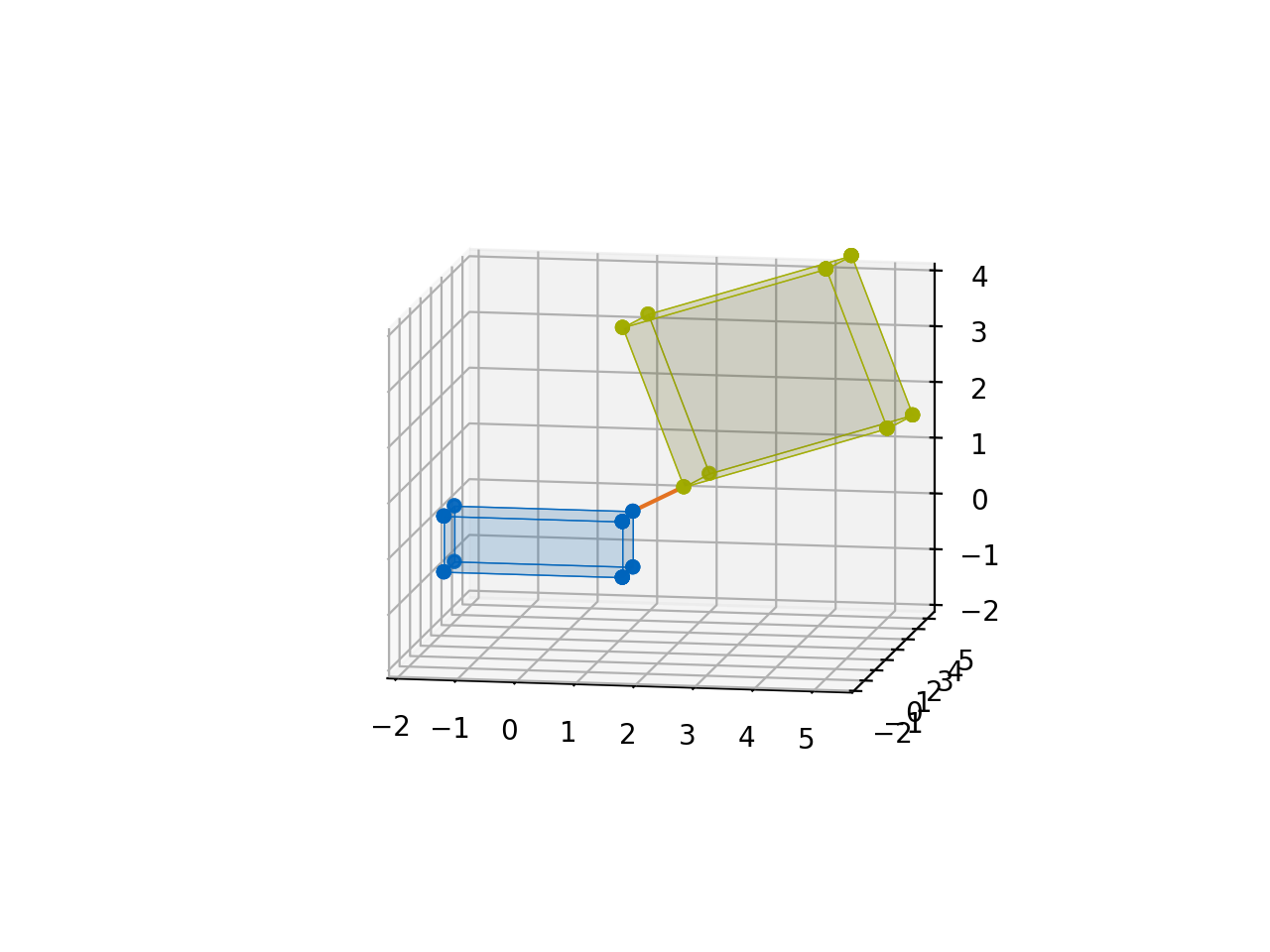}
        \caption{}
        \label{fig:c-c}
    \end{subfigure} 
    \caption{The different cases of v2v.}
    \label{fig:cases}
    \vspace{-0.7em}
\end{figure}

Similar to calculating the volumetric IoU, point-pairs of interest (PPOIs) can be defined, one of which gives the shortest distance $d_s$ between two bounding boxes $T^{1}$, $T^{2}$ such that
\begin{align}
    d_s = \textit{norm} ( P^{1}_s-P^{2}_s) < \textit{norm} ( P^{1}_i-P^{2}_j),&\\
            i \in \textit{hull}(T^1)\, \&\, j \in \textit{hull}(T^2)& \nonumber
\end{align}
Where $P^{1}_s$, $P^{2}_s$ denotes the two points from where the shortest distance is measured. Notice that the surface of the bounding boxes is an infinite, non discrete set of points, however, only a discrete set of points can be candidates for the pair of $P_s$.\\
The relevant point-pairs, shown in Fig~\ref{fig:cases}, are as follows

\begin{itemize}[leftmargin=1em]
	\itemsep-0.4em
    \item Corners of the first cuboid with their rectangular projection on the faces of the second cuboid \hyperref[fig:c-f]{(a)}
    \item Corners of the second cuboid with their rectangular projection on the faces of the first cuboid
    \item Corners of the first cuboid with their rectangular projection on the edges of the second cuboid \hyperref[fig:c-e]{(b)}
    \item Corners of the second cuboid with their rectangular projection on the edges of the first cuboid
    \item The distance defined by two edges, each of one cube \hyperref[fig:e-e]{(c)}
    \item The distance defined by two corners, each of one cube \hyperref[fig:c-c]{(d)}
\end{itemize}
In order to compute the point-pairs and the corresponding distances, we formulate equations for each point-face, point-edge, edge-edge projections.

\subsubsection{Point-Plane Projection}
The shortest distance $d$ between a point and a plane is given by its projection. In case of corners $l$ of one cuboid $i$ and the faces $k$ of another cuboid $j$, it is defined by
\begin{align}
    v &= C_{l}^i - f_{k, 0}^j \\
    d &= {n_{k}^j}^T v \\
    p_{\textit{prj}} &= C_{l}^i + n_{k}^j * d
\end{align}
In order to check if the projected point $p_{\textit{prj}}$ is within the face $f_k^j$ of the cube $j$  one can calculate the parameters $t_{p,1}$ and $t_{p,2}$ of its plane-function $P_k^j$. This can be done by calculating the pseudo-inverse of $N_k^j$.
\begin{equation}
    \begin{bmatrix}
        t_{p,1} \\
        t_{p,2}
    \end{bmatrix} = \left({N_k^j}^T N_k^j \right)^{-1} N_k^j \; v = {N_k^j}^+ \; v
\end{equation}
Only if both parameters are between $0$ and $1$, the projected point is a valid point.

\subsubsection{Point-Line projection}
The shortest distance $d$ between a point and a line is given by its projection. In case of corners $l$ of one cuboid $i$ and the edges $k$ of another cuboid $j$, it is defined by
\begin{align}
    t &= \frac{\left(C_{l}^i - l_{k, 0}^j\right)^T m_{k}^j}{{m_{k}^j}^2}  \\
    p_{\textit{prj}} &= L_{k}^j(t)
\end{align}
Only if parameter $t$ has a value between $0$ and $1$, the projected point is a valid point.
The solution is only defined for non-parallel lines, which in our case is sufficient. In parallel cases, the shortest distance is already given by the point to point or point to edge distance.

\subsubsection{Line-Line projection}
The shortest distance $d$ between a line and another line is given by its projection. In case of edges $k$ of one cuboid $i$ and the edges $l$ of another cuboid $j$, it is defined by
\begin{align}
    v &=  e_{k}^i - e_{l}^j  \\
    det &= {m_{k}^i}^2 + {m_{l}^j}^2 + {m_{k}^i}^T m_{l}^j \\
    \nonumber \\
    t^i &= \frac{{m_{l}^j}^2 \left({m_{k}^i}^T v\right) - \left({m_{l}^j}^T v\right) \left({m_{k}^i}^T m_{l}^j\right)}{det}
\end{align}
\begin{align}    
    t^j &= \frac{- {m_{k}^i}^2 \left({m_{l}^j}^T v\right) + \left({m_{k}^i}^T v\right) \left({m_{k}^i}^T m_{l}^j\right)}{det}\\
    \nonumber \\
    p_{\textit{prj}}^i &= L_{k}^i(t^i) \\
    p_{\textit{prj}}^j &= L_{l}^j(t^j)
\end{align}
The formula for the parameters $t^i,t^j$ can be derived by solving the minimization-problem $\left(L_{k}^i-L_{l}^j\right)^2$.

\subsubsection{Shortest Distance}
Every combination results in a maximum of $496$ point-pairs of interest. Since all functions only consist of linear equations, efficient methods of parallel computing can be applied. As a last step left, one must sort the list of candidates according to a vector-norm, such as L2 of the point-pair difference. Using L2 has the advantage that the values are already byproducts of the previously introduced point-to-plane projection. The shortest value of all PPOIs then corresponds to the shortest volume-to-volume distance.

\subsection{Bounding Box Disparity}
As a last metric we introduce the bounding box disparity (BBD), which is a combination of IoU and v2v. Since the Intersection over Union can only rank the similarity of two bounding boxes when they are overlapping, a distinction between two bounding boxes closer or farther away without an overlap cannot be made. Hence, we suggest the combination of IoU and v2v in the following way
\begin{equation}
    BBD = 1-IoU + v2v
\end{equation}
such that a continuous positive metric for the (dis-)similarity of two bounding boxes can be calculated. IoU can have values between $0$ and $1$, whereas $1$ corresponds to a total match and $0$ to no overlap. v2v on the other hand is $0$ as long as there is overlap and increases with further distance/mismatch of the bounding boxes. This results in a first quickly, but then linearly increasing scalar-field $BBD$.

\section{Conclusion}
With this paper we publish an open source library\footnote{\url{https://github.com/M-G-A/3D-Metrics}} for multiple 3D metrics for object detection, including the first analytic solution and its implementation of volumetric Intersection over Union and volume-to-volume distance for two bounding boxes with full degree of freedom. Further, we introduce the combined metric Bounding Box Disparity. In future work this could be extended such that for non overlapping bounding boxes also the rotation and size differences are considered.

\vfill\pagebreak

\bibliographystyle{IEEEbib}
\bibliography{bib}

\end{document}